\documentclass[10pt,twocolumn,letterpaper]{article}

\usepackage[pagenumbers]{cvpr} 
\usepackage[accsupp]{axessibility}  

\usepackage{graphicx}
\usepackage{amsmath}
\usepackage{amssymb}
\usepackage{booktabs}
\usepackage[ruled,vlined,linesnumbered]{algorithm2e}
\usepackage{color,colortbl}
\definecolor{LightCyan}{rgb}{0.88,1,1}
\definecolor{lightGreen}{rgb}{.935,.99,.935}
\definecolor{lightBlue}{rgb}{.935,.935,.99}
\usepackage[pagebackref,breaklinks,colorlinks]{hyperref}

\usepackage[capitalize]{cleveref}
\crefname{section}{Sec.}{Secs.}
\Crefname{section}{Section}{Sections}
\Crefname{table}{Table}{Tables}
\crefname{table}{Tab.}{Tabs.}

\def\cvprPaperID{10829} 
\def\confName{CVPR}
\def\confYear{2022}

\begin{document}

\title{Long-tailed Visual Recognition via Gaussian Clouded Logit Adjustment}
\author{
	Mengke Li$^{1}$ \quad Yiu-ming Cheung$^{1}$\thanks{Yiu-ming Cheung is the Corresponding Author.} \quad Yang Lu$^{2}$ \vspace{.3em}\\
	$^1$Department of Computer Science, Hong Kong Baptist University, Hong Kong\\
	$^2$Department of Computer Science and Technology, School of Informatics, Xiamen University, China\\
	{\tt\small \{csmkli, ymc\}@comp.hkbu.edu.hk, luyang@xmu.edu.cn}
}

\maketitle

\begin{abstract}
Long-tailed data is still a big challenge for deep neural networks, even though they have achieved great success on balanced data. We observe that vanilla training on long-tailed data with cross-entropy loss makes the instance-rich head classes severely squeeze the spatial distribution of the tail classes, which leads to difficulty in classifying tail class samples. Furthermore, the original cross-entropy loss can only propagate gradient short-lively because the gradient in softmax form rapidly approaches zero as the logit difference increases. This phenomenon is called softmax saturation. It is unfavorable for training on balanced data, but can be utilized to adjust the validity of the samples in long-tailed data, thereby solving the distorted embedding space of long-tailed problems. To this end, this paper proposes the Gaussian clouded logit adjustment by Gaussian perturbation of different class logits with varied amplitude. We define the amplitude of perturbation as cloud size and set relatively large cloud sizes to tail classes. The large cloud size can reduce the softmax saturation and thereby making tail class samples more active as well as enlarging the embedding space. To alleviate the bias in a classifier, we therefore propose the class-based effective number sampling strategy with classifier re-training. Extensive experiments on benchmark datasets validate the superior performance of the proposed method. Source code is available at \href{https://github.com/Keke921/GCLLoss}{\textcolor{blue}{https://github.com/Keke921/GCLLoss}}.
\end{abstract}

\section{Introduction} \label{sec:intro}
Deep neural networks (DNNs) have been widely utilized in a variety of visual recognition problems~\cite{he2016deep, he2017mask, ren2016faster, Wang2017NormFace} by virtue of the large-scale, high-quality, and annotated datasets. DNNs usually require the training dataset to be artificially balanced and have sufficient samples of each class. Unfortunately, from a practical perspective, object frequency usually follows a power law and typically exhibits a long-tailed distribution. Naive learning on such data is prone to undesirable bias towards the head classes which occupy the majority of the training samples~\cite{zhang2021survey}. Since tail classes have few training samples that cannot cover the real distribution in embedding space, their spatial span is severely compressed by head classes. In addition, a vast number of head class samples generate overwhelming discouraging gradients for tail classes. Thus, the learning of a classifier is biased towards the head classes. As a result, directly training on long-tailed data brings two key problems: 1) the distorted embedding space, and 2) the biased classifier.

\begin{figure}[t]
\centering
 \includegraphics[width=0.46\textwidth]{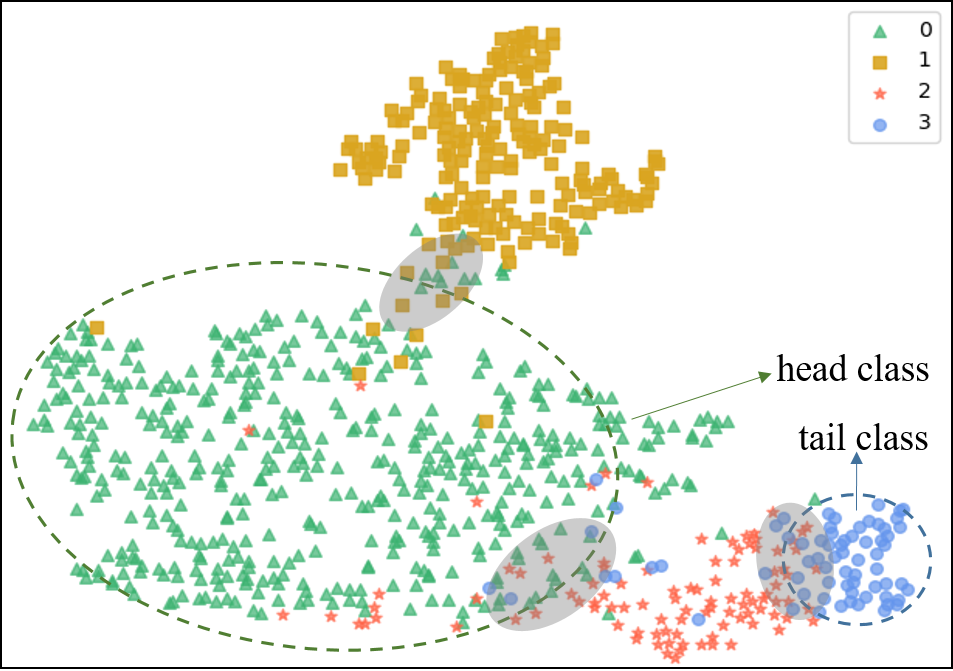}
 \caption{t-SNE visualization of the distorted embedding space. (Color for the best view.) The embeddings are calculated with ResNet-32 on a subset with four classes of CIFAR-10-LT. We randomly select four classes with the training numbers 500, 200, 100, and 50, respectively. The gray areas show the obscure regions between different classes.}
 \label{fig:intro-toy}
 \vspace{-6pt}
\end{figure}

In the literature, most of the recently proposed approaches focus on addressing the second problem only, \ie, the biased classifier. For example,  Menon \etal~\cite{adjustment21} and Hong \etal~\cite{Hong2021CVPR} applied post-adjust strategy to the trained model to calibrate the class boundary. Nevertheless, the distorted embedding cannot be adjusted with the post-hoc calibration, which is not conducive to further improving the model performance. Most recently, the two-stage decoupling methods~\cite{bbn20, decouple20, Kaidi2019, wang2020devil, DisAli21} have been proposed to obtain good embeddings in the first stage and then re-balance the classifier in the second stage. These methods obtain the representation by cross-entropy (CE) loss, which, however, leads to a severely uneven distributed embedding space. We implement a toy experiment to illustrate the distortion of the embedding space as shown in \cref{fig:intro-toy}, where t-SNE~\cite{van2008visualizing} is utilized to visualize the features of a long-tailed subset from CIFAR-10 dataset. We can observe that the tail class occupies a much small spatial span than the head class. This is because the tail class with fewer samples cannot cover the ground truth distribution. Moreover, \cref{fig:intro-toy} also shows that there are obscure regions (\ie, the grey area) between different classes. Softmax saturation~\cite{Chen2017CVPR} is one of the factors of these obscure regions because it leads to insufficient training. These obscure regions have a severe effect on the tail classes but little on the head classes. Since tail class samples clustered around the class boundary aggravate their spatial squeezing, while the head class samples with enough variety can already cover the true distribution.

\begin{figure}[t]
\centering
 \includegraphics[width=0.5\textwidth]{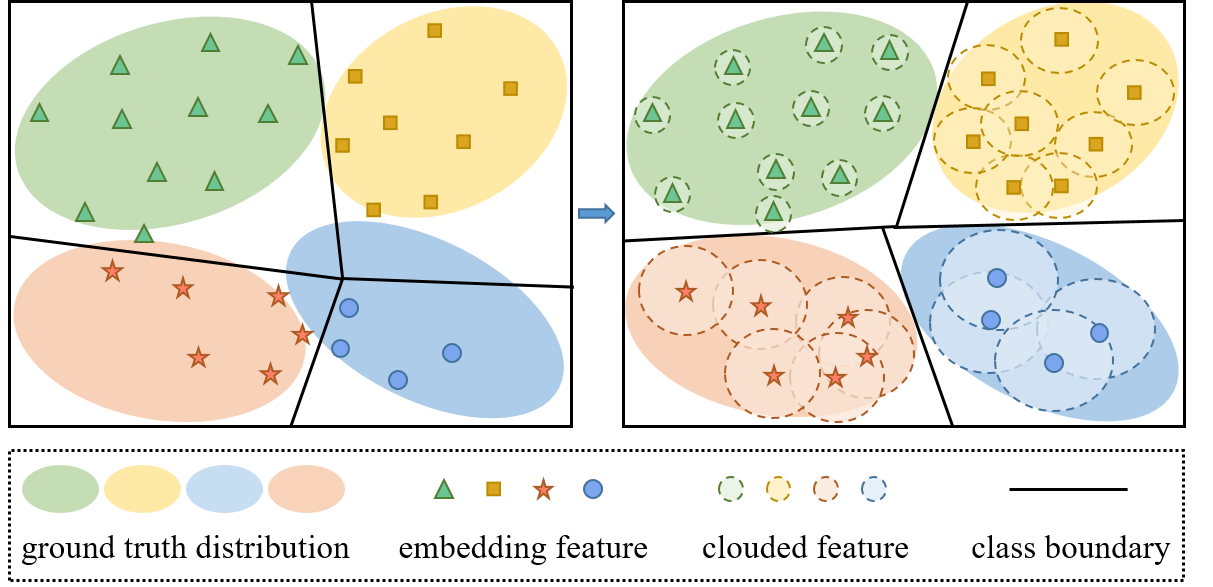}
 \caption{An overview of GCL. (Color for the best view.) The tail class logit is assigned to a larger sample cloud size than the head class, which corresponds to a large relative cloud size of the feature in the direction of the tail class anchor. In this way, the distortion of the embedding space can be calibrated well.}
 \label{fig:intro}
 \vspace{-6pt}
\end{figure}

Softmax saturation refers to the inopportune early gradients vanishing produced by the softmax~\cite{Chen2017CVPR, Zhang2021class}, which weakens the validity of training samples and impedes model training. However, from another perspective, the seemingly harmful softmax saturation has the ability to balance the valid samples of different classes and thus help calibrate the distortion of embedding space. Specifically, we disturb the logit of different classes with different amplitudes. We name the disturbed logit as Gaussian clouded logit (GCL) and the amplitude of the disturbance as cloud size, because we set the disturbance to a Gaussian distribution. The tail classes have few training samples and thus the training samples of them should be more valid. We therefore disturb the logit of tail classes with large relative cloud sizes to reduce the softmax saturation. In this way, tail class samples can provide more gradients without overfitting and thus indirectly affect their embedding space. In addition, a large cloud size of the tail class logit corresponds to the large cloud size on feature in the direction of the class anchor. Therefore, tail classes can have large margins towards the class boundary, so as to alleviate the severe uneven distribution between the head and tail classes. Conversely, the head classes are set to small cloud sizes, so that they can be automatically filtered out during training. Eventually, as shown in ~\cref{fig:intro}, the tail class samples can be pushed more away from the class boundary so as the distortion of the embedding space can be calibrated.

To address the biased classifier, we re-balance the training data with a class-wise sampling strategy. As training with GCL makes the validity of different classes vary, the so-called ``effectiveness''~\cite{cui2019class} of them are different. Existing class-wise balanced sampling strategies will lead to excessive training of tail classes for GCL. We thereby propose the class-based effective number (CBEN) sampling strategy, which is based on sample validity and label frequencies to re-balance the classifier. This simple but effective sampling strategy helps mitigate the classifier bias towards the head classes and further boost the performance of GCL.

Extensive experiments on multiple commonly used long-tailed recognition benchmark datasets demonstrate that the proposed GCL surpasses the recently proposed counterparts. In summary, the key contributions of our work are three-fold:
\begin{itemize}
    \item We propose the GCL adjustment loss function, which utilizes softmax saturation to balance the sample validity of different classes. An evenly distributed embedding can be obtained with the proposed GCL.
    \item We propose a simple but effective class-based effective number (CBEN) sampling strategy for re-balancing the classifier to avoid repeat training of tail classes. This sampling strategy can further boost the performance of GCL.
    \item Extensive experiments on popular long-tailed datasets demonstrate that the proposed method outperforms the state-of-the-art counterparts.
\end{itemize}


\section{Related Works}
\label{sec:related_works}
Long-tailed classification is one of the long-standing research problems in machine learning. Several kinds of approaches have been proposed to address it. This section briefly introduces the most related three regimes, namely loss modification, logit adjustment, and decoupling representation.
\subsection{Loss Modification}
Modifying the loss function through re-weighting is the most natural method. Sample-wise re-weighting methods~\cite{Mengye2018Learning, Tsung2020Focal} attempt to make the model pay more attention to the difficult samples by introducing fine-gained coefficients in the loss for imbalanced learning. For example, focal loss~\cite{Tsung2020Focal} introduces a tunable focusing parameter, which is negatively correlated with the predicted probability of the target class. This focusing parameter helps the model training focus on hard samples and prevents the numerous easy negatives from overwhelming. Class-wise re-weighting methods~\cite{Huang2016CVPR, Salman2018Cost, cui2019class, tan2020Equalization} assign the standard CE loss with category-specific parameters that are inversely proportional to the class frequencies. For example, Tan \etal~\cite{tan2020Equalization} proposed equalization loss, which utilizes a weight term to randomly ignore the discouraging gradients of head class samples. These methods can alleviate the data imbalance to a certain extent. However, the classification difficulty of a sample is not directly related to its corresponding class size. Further, another side effect of assigning higher weights to difficult samples/tail classes is overly focusing on harmful samples (\eg, noisy data or mislabeled data)~\cite{Pang2017Understanding}.

\subsection{Logit Adjustment}
Logit adjustment assigns relatively large margins for tail classes. Most recently, Menon \etal~\cite{adjustment21} have proposed a logit adjustment (LA) method which is consistent with minimizing the balanced error. The logit shifting in LA of different classes is based on label frequencies of training data. Differently, LADE~\cite{Hong2021CVPR} calibrates the logit to the test set using the label distribution of test data, so that the test set can also be imbalanced. Tang~\etal~\cite{De-confound-TDE20} adopted causal intervention to remove the ``bad'' SGD momentum and keep the ``good'' one to avoid the harmful causal effect for tail prediction. DisAlign~\cite{DisAli21} adjusts the logit by calibrating the model prediction to a reference distribution of classes that favors the balanced prediction. These methods well adjust the model logits through post-hoc shifting but without considering calibrating the embedding space. Another type of approach~\cite{Kaidi2019, Cao2020CVPR} addresses long-tailed data by leaving large relative margins for tail classes during training. For example, label-distribution-aware margin (LDAM) loss proposed by Cao~\etal~\cite{Kaidi2019} utilizes Rademacher complexity to theoretically prove that the margin should be inversely proportional to a quarter power of label frequencies. The hard margin on target logit helps make the intro-class samples more compact, but does not truly enlarge the tail class span in embedding space.

\subsection{Decoupling Representation }
Many recent works have focused on improving the long-tailed visual recognition performance by decoupling the representation and classifier.
Most recently, LDAM-DRW~\cite{Kaidi2019} has been proposed, which learns features in the first stage and adopts the deferred re-weighting (DRW) to fine-tune the decision boundary in the second stage. It significantly improves the long-tailed prediction accuracy, but the theoretical explanation of DRW is not clear. After that, Kang~\etal~\cite{decouple20} precisely pointed out that the learning process of representation and classifier can be decoupled into two separate stages. The representation learning is conducted on the original long-tailed data in the first stage and the classifier learning is performed on class-balanced re-sampling data in the second stage. A lot of works \cite{wang2020devil, Wang2020Frustratingly, DisAli21, mislas21} have further refined this strategy. For example, Zhang~\etal~\cite{DisAli21} proposed an adaptive calibration function to calibrate the predicted logits of different classes into a balanced class prior in the second stage. Zhong~\etal~\cite{mislas21} proposed label distribution-based soft label to deal with different degrees of over-confidence for classes and can improve the classifier learning in the second stage. Another alternative direction is proposed by Zhou~\etal~\cite{bbn20}, which splits the network structure into two branches that focus on learning the representation of head and tail classes, respectively. This method incorporates feature mixup~\cite{verma2019manifold} into a cumulative learning strategy and also achieves the state-of-the-art results. Following~\cite{bbn20}, Wang~\etal~\cite{contrastive21} introduced contrastive learning into this bilateral-branch network to further improve the long-tailed classification performance.

\section{Proposed Approach: GCL}\label{method}
The key idea of our proposed GCL is to utilize the softmax saturation to automatically balance the valid samples of head and tail classes. The theoretical motivation and the formulation of the loss function of the proposed approach are presented as follows.

\subsection{Motivation}\label{sec:motivation}

\cref{fig:intro-toy} shows that the obscure region among different classes, especially the tail class, is large. One important factor of this obscure region is the softmax saturation in CE loss~\cite{Chen2017CVPR}. Suppose $\{x,y\}\in \mathcal{T}$ represents a sample $\{x,y\}$ from the training set $\mathcal{T}$ with the total $N$ samples in $C$ classes, and $y \in \{1,\ldots, C\}$ is the ground truth label. The softmax loss function for the input image $x$ can be written as:
\begin{equation}\label{eq:softmax}
  \mathcal{L}(x) = - \log p_y, \text{ with } p_y = \frac{e^{z_{y}}}{\sum_{j=1}^C e^{z_j}},
\end{equation}
where $z_j$ represents the predicted logit of class $j$. We use the subscript $y$ $(j\neq y)$ to represent the target class. That is, $z_{y}$ indicates the target logit and $z_{j} (j\neq y)$ is the non-target logit.

In backward propagation, the gradients on $z_j$ is calculated by:
\begin{equation}\label{eq:partial_sm}
\frac{\partial \mathcal{L}}{\partial z_j} =
\left\{
\begin{array}{lr}
p_j-1, &j = y\\
p_j,   &j\neq y.
\end{array}
\right.
\end{equation}
Without loss of generality, we use the binary classification as an example. Supposing $x$ is from class 1, the gradients on $z_1$ is then calculated by:
\begin{equation}\label{eq:bi_partial}
\frac{\partial \mathcal{L}}{\partial z_1} = -\frac{1}{1+e^{z_1-z_2}}.
\end{equation}
\cref{eq:bi_partial} indicates that the gradient of the target class rapidly approaches zero with the increase of the logit difference. Softmax can only slightly separate various classes, and lacks the power to evenly distribute each class in the embedded space. Therefore, there are many overlapping areas among the classes. In particular, under the circumstances of long-tailed classification, the tail class features are not sufficient to cover the real distribution in embedding space. The early gradient vanish caused by softmax saturation exacerbates the squeezing of their embedding space. A straightforward approach is to introduce hard margin~\cite{Deng2019ArcFace, Kaidi2019, Zhang2021class}. However, the hard margin will cause the samples to shrink towards the class anchor and easy to overfit tail classes, which cannot evenly distribute the embedding space well. Fortunately, softmax saturation can help filter out the head class samples and make the tail class samples fully participate in training. In this way, the tail classes can be pushed away from the head classes and indirectly enlarge their embedding space. 

\subsection{Embedding Space Calibration}\label{sec:sc}
Suppose the features of different class samples satisfy Gaussian distribution. We can obtain a disturbed feature $\mathbf{f}^{cld}$ of the input by Gaussian sampling, which is represented as:
\begin{equation}
    \mathbf{f}^{cld} \triangleq  \mathbf{f}+\delta\mathbf{E},
\end{equation}
where $\mathbf{f}\in\mathbb{R}^D$ is the feature obtained from the embedding layer with the dimension of $D$. $\mathbf{E}\sim~\mathcal{N}(\mathbf{u},\mathbf{\Sigma})$ is the disturbance sampled from Gaussian distribution, and the mean vector and covariance matrix are represented by $\mathbf{u}\in\mathbb{R}^D$ and  $\mathbf{\Sigma}\in\mathbb{R}^{D\times~D}$, respectively. $\delta>0$ is a parameter that is used to adjust the amplitude of disturbance. In addition, $\delta$ should be a small number because a large disturbance will mislead the model. This disturbed feature is the input of the classifier. We use $\mathbf{W}=\{\mathbf{w}_1, \mathbf{w}_2,\cdots,\mathbf{w}_C\} \in \mathbb{R}^{D\times C}$ to represent the weight matrix of the classifier, where $\mathbf{w}_j$ represents the anchor vector of class $j$ in the classifier. Then, the corresponding disturbed logit $z^{cld}_j$ of class $j$ is calculated by:
\begin{equation}
\begin{array}{lll}
  z^{cld}_j &= \mathbf{w}^T_{j}\mathbf{f}^{cld} + \mathbf{b}_{j}\\
  \\[-6pt]
  &= \mathbf{w}^T_{j}\mathbf{f} + \mathbf{b}_{j} + \mathbf{w}^T_{j}(\mathbf{\delta\mathbf{E}})\\
  \\[-6pt]
  &= z_{j} + \delta(\mathbf{w}^T_{j} \mathbf{E}). \\
\end{array}
\end{equation}
As the range of $z^{cld}_j$ is enlarged with random Gaussian disturbances, we call it Gaussian clouded logit, and $\delta(\mathbf{w}^T_{j} \mathbf{E})$ is the clouded term. Please note that the clouded term has the different degrees of influence on the final predicted results based on different predicted logits. It has a relatively small impact on $z^{cld}_j$ when the original logit $z_j$ is large. On the contrary, it will play a key role for $z^{cld}_j$ when $z_j$ is small. As a result, we need to normalize the effect caused by different predicted logits and maintain the consistency of the influence of the clouded term. Inspired by \cite{Wang2017NormFace, wang2018cosface, Deng2019ArcFace}, we normalize the clouded logits based on cosine distance. In this way, the norm of the feature and the class anchor can be normalized to the fixed numbers. We use $s_1$ and $s_2$ to represent these two numbers. The normalized clouded logit is named \textit{clouded cosine logit}, which is calculated by:

\begin{equation}\label{eq:cld_nor_scr}
\begin{array}{ll}
  \Tilde{z}^{cld}_{j}&=\displaystyle \frac{s_1\mathbf{w}^T_{j}\cdot s_2\mathbf{f}^{cld}}{\|\mathbf{w}^T_{j}\|\|\mathbf{f}^{cld}\|} \\
  \\[-6pt]
  & = \displaystyle s\cdot(\frac{\mathbf{w}^T_{j}\mathbf{f}}{\|\mathbf{w}^T_{j}\|\| \mathbf{f}+\delta\mathbf{E}\|} + \delta\frac{\mathbf{w}^T_{j}\mathbf{E}}{\|\mathbf{w}^T_{j}\|\| \mathbf{f}+\delta\mathbf{E}\|})\\
\end{array},
\end{equation}
where $s = s_1 \cdot s_2$ is a constant. In the first term of \cref{eq:cld_nor_scr}, $\| \mathbf{f}+\delta\mathbf{E}\| \approx \| \mathbf{f} \| $ because $\delta$ is a small number. In the second term, the norm of feature is normalized to $s_1$. Thus, $\Tilde{z}^{cld}_{j}$ can be simplified as:
\begin{equation}
  \Tilde{z}^{cld}_{j} \approx s\cdot(\frac{\mathbf{w}^T_{j}\mathbf{f}}{\|\mathbf{w}^T_{j}\|\| \mathbf{f}\|} + \frac{\delta}{s_1} I_j\mathbf{E}),
\end{equation}
where $I_j$ is the identity vector that has the same direction as $\mathbf{w}^T_j$. In order to simplify the calculation, we make the clouded cosine logit still satisfy the Gaussian distribution. Thus, we introduce a constant $\sigma$ and set the covariance matrix $\mathbf{\Sigma} = \sigma\mathbf{I}$, where $\mathbf{I} \in \mathcal{R}^{D\times~D}$ is the identity matrix. Then, $I_j\mathbf{E}$ is the projection of the noise sampled by Gaussian in the direction of the anchor vector of class $j$. We denote its magnitude by $\varepsilon_j$. Therefore, $\Tilde{z}^{cld}_{j}$ can be calculated by:
\begin{equation}\label{cld_z_tmp}
\begin{array}{cc}
\Tilde{z}^{cld}_{j} & = s\cdot(\Tilde{z}_{j} + \frac{\delta}{s_1}\varepsilon_j)  \\
     \\[-6pt]
     &\Leftrightarrow s\cdot( \Tilde{z}_{j} + \delta_j\varepsilon)
\end{array},
\end{equation}
where $\Tilde{z}_j = \cos \theta_j$ is the cosine distance, and $\theta_j$ is the angle between $\mathbf{f}$ and $\mathbf{w}_j$. $\delta_j$ is the logit cloud size that depends on different classes.

To achieve the two goals mentioned in \cref{sec:motivation}, \ie, 1) encourage tail class samples to participate more in training; 2) enlarge the embedding space for the tail classes, the size of logit cloud should be negatively correlated with the number of training samples. For the most frequent class, the diversity of training samples is sufficient and we set its logit cloud size to zero, while utilizing larger cloud sizes for tail classes. The merits of this large relative cloud size of tail classes are three-fold: 1) reduce the softmax saturation and thereby increase the training degree of tail classes; 2) different values are sampled randomly from the Gaussian cloud so as to avoid overfitting; 3) enlarge the margin of class boundary for tail classes and can calibrate the distortion of the embedding space. We therefore empirically set the cloud size for class $j$ as:
\begin{equation}\label{cloud_size}
\delta_j = \log n_{max}-\log n_j,
\end{equation}
where $n_{max}$ is the sample numbers of the most frequent class. We experimentally verify the effectiveness of this cloud size adjustment strategy in \cref{sec:cloud_size} .

The Gaussian clouded logit difference $\Delta_{y\_j}$ between the target and non-target classes is:
\begin{equation}
\begin{array}{ll}
\Delta_{y\_j} & =z^{cld}_y-z^{cld}_j\\
\\[-6pt]
& = z_y-z_j+\varepsilon ({\delta_y-\delta_j})
\end{array}.
\end{equation}
If $\varepsilon >0$, $\Delta_{y\_j}$ for tail classes will be increased. However, our goal is to reduce the logit difference to alleviate the softmax saturation for tail classes. In addition, a reduced logit corresponds to the feature that is relatively far from the class anchor. If the relatively distant feature can be predicted correctly, the closer one will be able to assign the right label. Therefore, we require $\varepsilon$ to be negative. Subsequently, the clouded cosine logit can be written in the following form:
\begin{equation}\label{eq:z_cld_i}
\Tilde{z}^{cld}_j=  s\cdot(\Tilde{z}_j  - \delta_j\|\varepsilon\|).
\end{equation}

Taking the clouded cosine logit into the original softmax, we can obtain the loss function of GCL:
\begin{equation}\label{gcl}
  \mathcal{L}_{GCL} = -\frac{1}{N}\sum_i\log \frac{e^{\Tilde{z}^{cld}_{y_i}}}{\sum_j e^{\Tilde{z}^{cld}_j}}.
\end{equation}

\subsection{Classifier Re-balance}\label{sec:re-balance-cls}
The gradients derived in \cref{eq:partial_sm} demonstrate that the sample of the target class $y$ punishes the classifier weights $\textbf{w}_j$ of non-target class $j, j\neq y$ w.r.t. $p_j$. The head classes have enormously greater training instances than tail classes. Therefore, the classifier weights of tail classes receive much more penalty than positive signals during training. Consequently, the classifier will bias towards the head classes, and the predicted logits of the tail classes will be seriously suppressed, resulting in low classification accuracy of the tail classes. A straightforward approach is to use the re-sampled data to re-train the classifier. We apply the classifier re-training (cRT), which was adopted by Kang~\etal~\cite{decouple20} and Wang~\etal~\cite{wang2020devil}. As the GCL loss enables different class samples to participate in training to different degrees, the effectiveness of different class samples is varied. Class-balanced sampling will lead to repeat training for tail classes. Drawing on the effective number proposed by Cui \etal~\cite{cui2019class}, we propose the class-based effective number (CBEN) sampling to avoid excessive training of tail classes. The sampling probability $\rho_j$ of a sample from class $j$ is calculated by:
\begin{equation}\label{eq:sp}
  \rho_j= \frac{1-\beta_j}{1-\beta_j^{n_j}}.
\end{equation}
Since the sum of the sampling probability for all data needs to be 1, we normalize $\rho_j$ by $\rho_j \leftarrow \frac{\rho_j}{\sum_i \rho_i}$. $\beta_j$ reflects the validity of different class samples. The class samples with large cloud size participate more in training. Therefore, $\beta_j$ is positively correlated with cloud size $\delta_j$. We set $\beta_j$ as:
\begin{equation}\label{eq:beta}
  \beta_j= b \times \frac{\delta_j-\delta_{min}}{\delta_{max}-\delta_{min}}+a,
\end{equation}
so that $\beta_j$ can be in the region $[a,a+b]$, where $a$ and $b$ are the range hyper-parameters.

The overall training procedure of the proposed method is summarized in \cref{algorithm}.

\begin{algorithm}[t]
\caption{Gaussian clouded logit}\label{algorithm}
\SetAlgoLined
\KwIn {Training dataset $\mathcal{T}$\;}
\KwOut {Predicted labels\;}
Initialize the model parameters $\omega$ of the CNN network $\phi((x,y);\omega)$\ randomly \;
\For {$iter=1$ to $I_0$}{
Sample a batch samples $\mathcal{B}$ from the original long-tailed data $\mathcal{T}$ with batch size $b$\;
Obtain the logit cloud size:
$\delta_j \leftarrow  \log n_{max}-\log n_j$\;
Calculate the loss by \cref{gcl}:
$\:\:\:\: \mathcal{L}((x,y);\omega) = \frac{1}{b}\sum_{(x,y)\in \mathcal{B}} \mathcal{L}_{GCL}(x,y)$\;
Update model parameters:
$\:\:\:\: \omega = \omega - \alpha \nabla_{\omega} \mathcal{L}((x,y);\omega)$.
}
\For{$iter=I_0 + 1 $ to $I_0+I_1$}{
Calculate sampling rate:
$\beta_j \leftarrow b \times \frac{\delta_j-\delta_{max}}{\delta_{max}-\delta_{min}}+a$;
$\rho_j \leftarrow \frac{1-\beta_j^{n_j}}{1-\beta_j}$;
$\rho_j \leftarrow \frac{\rho_j}{\sum_i \rho_i}$\;
Sample a batch samples $\mathcal{B'}$ with the sampling probability $\rho_j$ and the batch size $b$\;
Calculate the loss by \cref{gcl}:
$\mathcal{L}((x,y);\omega) = \frac{1}{b}\sum_{(x,y)\in \mathcal{B'}} \mathcal{L}_{GCL}(x,y)$\;
Update classifier parameters $\omega_{cls}$ (representation parameters are frozen):
$\omega_{cls} = \omega_{cls} - \alpha \nabla_{\omega_{cls}} \mathcal{L}((x,y);\omega_{cls})$.
}
\end{algorithm}

\section{Experiments}\label{sec:exp}
\subsection{Datasets}
We use five benchmarks: long-tailed CIFAR datasets that include CIFAR-10-LT and CIFAR-100-LT, long-tailed ImageNet-2012 (ImageNet-LT), iNaturalist 2018~\cite{Horn_2018_CVPR} and long-tailed Places-2 (Places-LT). The original version of CIFAR-10/100~\cite{krizhevsky2009learning}, ImageNet-2012~\cite{ILSVRC15} and Places-2~\cite{zhou2017places} are all balanced datasets. We follow Cao~\etal~\cite{Kaidi2019} and Cui~\etal~\cite{cui2019class} to create long-tailed versions of CIFAR-10/100 and use the long-tailed versions of ImageNet-2012 and Places-2 produced by Liu~\etal~\cite{OLTR19}.

\textbf{CIFAR-10/100-LT.}
The original CIFAR-10 and CIFAR-100 consist of 10 and 100 classes, respectively. They both have 60,000 color images of size $32\times32$. 50,000 of them are used for training and the remaining images are for validation. Following~\cite{cui2019class, Kaidi2019}, we down-sampling training samples per class with the exponential function $n_i = n_{o_i} \times \mu^i$, where $i$ is the class index (0-indexed), $n_{o_i}$ is the number of training samples in original CIFAR and $\mu \in (0,1)$. The validation sets are kept unchanged. The imbalance ratio $\gamma$ is defined as the ratio of the sample size of the most and the least frequent classes, \ie $\gamma=\max{(n_i)}/ \min{(n_i)}, i = 0,1,...,C-1$.  $\gamma$ is set at its common values, i.e. $\gamma = 50, 100$ and $200$, in our experiments.

\textbf{ImageNet-LT} and \textbf{Places-LT.}
The balanced versions of ImageNet-2012 and Places-2 are large-scale real-world datasets for classification and localization. We follow Liu \etal's work \cite{OLTR19} to construct the long-tailed version of these two datasets by truncating a subset with the Pareto distribution with the power value $\alpha=6$ from the balanced versions. The original balanced validation sets remain unchanged. Overall, ImageNet-LT has 115.8K training images from 1,000 categories with $\gamma=1,280/5$. Places-LT contains 62.5K training images from 365 categories with $\gamma=4,980/5$.

\textbf{iNaturalist 2018.}
The 2018 version of iNaturalist is a real-world fine-grained dataset for classification and detection, which exhibits extremely imbalanced distribution. It contains 437.5K training images and 24.4K validation images from 8,142 categories. We follow the official splits of training and validation sets in the experiments.

\subsection{Experimental Setting}
The pre-setting parameters in the first stage were the Gaussian distribution parameters $(\mu$, $\sigma^2)$ and the region $[a, b]$ of sample validity $\beta_j$.
We know that $\Tilde{z}_i \in [-1,1]$, thus the maximum feature cloud size cannot exceed 1. Since Gaussian distribution has a probability of about $99.7\%$ falling in $[\mu-3\sigma, \mu+3\sigma]$, we set $\mu=0$ and $\sigma=\frac{1}{3}$. We further clamped the $\varepsilon$ to $[-1,1]$ to prevent its amplitude from exceeding 1. We set $\beta_j \in [0.999, 0.9999]$, i.e. $a=0.999$ and $b=0.0009$. Moreover, we normalized $\delta_i, i = \{1,2,\cdots ,C\}$ by $\delta_i \triangleq \delta_i / \delta_{max}$ to ensure that the maximum value of $\delta_i$ did not exceed 1. Similar with Zhong~\etal~\cite{mislas21}, the mixup~\cite{Hongyi2018} strategy was also adopted in our experiments.

We utilized PyTorch~\cite{paszke2019pytorch} to implement all the backbones. SGD optimizer with momentum of 0.9 and the multi-step learning rate schedule were adopted. All the models were trained from scratch except ResNet-152 that was pre-trained on the original balanced version of ImageNet-2012. For the first stage, we selected ResNet-32 as the backbone network and followed the setting in Cao~\etal~\cite{Kaidi2019} for CIFAR-10/100-LT. For the large-scale dataset, namely ImageNet-LT, iNaturalist 2018, and Places-LT, we mainly followed Kang~\etal~\cite{decouple20} except the learning rate schedule. For the second stage, \ie, re-balancing the classifier, we followed Kang~\etal~\cite{decouple20} for all datasets.

\begin{table*}[t]
\renewcommand{\thefootnote}{\fnsymbol{footnote}}
 \centering  
 \caption{Comparison results on CIFAR-10/100-LT in terms of top-1 accuracy (\%), where the best and the second-best results are shown in \underline{\textbf{underline bold}} and \textbf{bold}, respectively. *indicates that the results are quoted from the corresponding references. The other results are obtained by re-implementing with the official codes.} \label{cifar_results}
 \setlength{\tabcolsep}{16pt}
 \renewcommand{\arraystretch}{0.88}
  {\begin{tabular}{c |c c c | c c c}
  \toprule[0.8pt]
  Dataset & \multicolumn{3}{c|}{CIFAR-10-LT} & \multicolumn{3}{c}{CIFAR-100-LT}\\ \hline
  \hline
  Backbone Net& \multicolumn{6}{c}{ResNet-32}\\ %
  \hline		
  Imbalance ratio &200 &100 &50  &200 &100 &50 \\
  \hline
  CE loss &65.68  &70.70  &74.81  &34.84 &38.43 &43.9\\
  CE loss + mixup~\cite{Hongyi2018} (2018)  &65.84 &72.96 &79.48 &35.84 &40.01 &45.16 \\
  \hline                                                
  LDAM-DRW \cite{Kaidi2019} (2019)& 73.52  &77.03  &81.03 &38.91 &42.04 &47.62\\
  De-confound-TDE \footnotemark[1] \cite{De-confound-TDE20} (2020)   &-  &80.60  &83.60 &- &44.15 &50.31\\
  CE loss + mixup + cRT~\cite{decouple20} (2020) & 73.06 &79.15 &84.21 & 41.73 & 45.12 & 50.86\\
  BBN~\cite{bbn20} (2020) &73.47 &79.82 &81.18 &37.21 &42.56 &47.02\\
  Contrastive learning \footnotemark[1]~\cite{contrastive21}  (2021) &- &81.40 & \textbf{85.36} &- &46.72 &51.87\\
  MisLAS~\cite{mislas21} (2021)   &\textbf{77.31}& \textbf{82.06} &85.16 &\textbf{42.33} &\textbf{47.50} &\textbf{52.62} \\
  \hline
  \multicolumn{1}{>{\columncolor{lightGreen}[16pt][344pt]}c|}
  {GCL} &\underline{\textbf{79.03}} &\underline{\textbf{82.68}} &\underline{\textbf{85.46}} &\underline{\textbf{44.88}} &\underline{\textbf{48.71}} &\underline{\textbf{53.55}} \\
  \bottomrule[0.8pt]
 \end{tabular}}
\end{table*}

\begin{table*}[t]
\renewcommand{\thefootnote}{\fnsymbol{footnote}}
 \centering  
 \caption{Comparison results on ImageNet-LT, iNaturalist 2018 and Places-LT in terms of top-1 accuracy (\%), where the best and the second-best results are shown in \underline{\textbf{underline bold}} and \textbf{bold}, respectively. *indicates that the results are quoted from the corresponding references. The other results are obtained by re-implementing with the official codes.} \label{large_dataset_results}
 \setlength{\tabcolsep}{18.8pt}
 \renewcommand{\arraystretch}{0.9}
  {\begin{tabular}{c |c|c|c }
  \toprule[0.8pt]
  Dataset & ImageNet-LT & iNaturalist 2018 & Places-LT\\ \hline
  \hline
  Backbone Net & ResNet-50 & ResNet-50  & ResNet-152  \\ %
  \hline
  CE loss  &44.51 &63.80   & 27.13 \\
  CE loss + mixup \cite{Hongyi2018} (2018)  &45.66 &65.77 &29.51 \\
  \hline
  LDAM-DRW \cite{Kaidi2019} \footnotemark[1] (2019) &48.80  &68.00 &-  \\
  OLTR~\footnotemark[1]~\cite{OLTR19} (2019) &-  &- &35.9  \\
  Decoupling~\cite{decouple20} (2020) &47.70  &69.49 &37.62  \\
  CE loss + mixup + cRT~\cite{decouple20} (2020) &51.68 &70.16 &38.51 \\
  Logit adjustment \footnotemark[1]\cite{adjustment21}(2021) &51.11 & 66.36 &- \\
  DisAlign  \footnotemark[1]\cite{DisAli21} (2021) &\textbf{52.91} &70.06 &39.30 \\
  MisLAS \cite{mislas21} (2021)   &52.11  &\textbf{71.57} & \textbf{40.15}\\
  \hline
  \multicolumn{1}{>{\columncolor{lightGreen}[18pt][305pt]}c|}
  {GCL} &\underline{\textbf{54.88}} &\underline{\textbf{72.01}} &\underline{\textbf{40.64}} \\
  \bottomrule[0.8pt]
 \end{tabular}}
 \vspace{-6pt}
\end{table*}
\subsection{Competing Methods}
To verify the effectiveness of the proposed method, we have conducted extensive experiments to compare with the previous methods, including the following two groups:

\textbf{Baseline Methods}.
We implemented vanilla training with cross-entropy (CE) loss as one of our baseline methods. Many visual recognition works~\cite{Pang2020Mixup, Zhang2021How, Kim2021Co-Mixup, zhang2021bag} have shown the efficacy of mixup, CE loss cooperated with mixup was therefore also compared.

\textbf{State-of-the-art Methods}.
The recently proposed representation learning method, namely OLTR~\cite{OLTR19} and logit adjustment method, namely De-confound-TDE inference~\cite{De-confound-TDE20} were compared. We also compared with the two-stage methods including LDAM-DRW~\cite{Kaidi2019} and MisLAS~\cite{mislas21}, which both achieve satisfactory classification accuracy on the aforementioned long-tailed datasets. For CIFAR-10/100-LT datasets, we made comparison with BBN~\cite{bbn20} and contrastive learning~\cite{contrastive21}. For the large-scale datasets, we compared with the most recently proposed two-stage methods, including decoupling~\cite{decouple20}, logit adjustment~\cite{adjustment21} and DisAlign~\cite{DisAli21}. For a fair comparison, we additionally conducted the comparison experiment with the two-stage strategy which added classifier re-training (cRT)~\cite{decouple20} to CE loss + mixup on all datasets.

\subsection{Comparison Results}
Comparative studies have been conducted to show the efficacy of the proposed GCL. The results are presented in \cref{cifar_results} and \cref{large_dataset_results}. We use top-1 accuracy on test sets as the performance metric. For the results from those papers that have yet to release the code or relevant hyper-parameters, we directly quote their results from the original papers.

\subsubsection{Experimental Results on CIFAR-10/100-LT}
The results on CIFAR-10/100-LT datasets are summarized in \cref{cifar_results}. We can observe that our proposed GCL outperforms the previous methods by notable margins with all imbalanced ratios. Especially for the largest one, \ie, $\gamma = 200$, the proposed approach has obvious improvement. We get $79.03\%$ and $44.88\%$ in top-1 classification accuracy for CIFAR-10-LT and CIFAR-100-LT with $\gamma = 200$, which surpasses the second best method, \ie, MisLAS by a significant margin of $1.72\%$ and $2.55\%$, respectively.

\subsubsection{Experimental Results on Large-scale Latasets}
The results on three large-scale long-tailed datasets, \ie, ImageNet-LT, iNaturalist 2018, and Place-LT, are reported in \cref{large_dataset_results}. Our approach is superior to prior art on all datasets. On ImageNet-LT, our method achieves $54.88\%$ top-1 accuracy, outperforming DisAlign by a large margin at $1.97\%$ and MisLAS at $2.77\%$, respectively. On iNaturalist 2018, the proposed approach achieves $72.01\%$ top-1 accuracy, which outperforms the second-best method by $0.44\%$. On Place-LT, our method achieves $40.64\%$ top-1 classification accuracy, with a performance gain at $0.49\%$ over MisLAS. Although the performance gain compared with MisLAS on iNaturalist 2018 and Place-LT is not as high as other datasets, our method does not require hyper-parameters searching for different datasets, and thus it is relatively easy to implement.

\subsection{Model Validation and Analysis}
We conduct a series of ablation studies to further analyze the proposed method.
\subsubsection{The Role of Gaussian Clouded Logit}
\begin{figure}
  \centering
  \subfloat[On training set]{
        \includegraphics[width=0.48\textwidth]{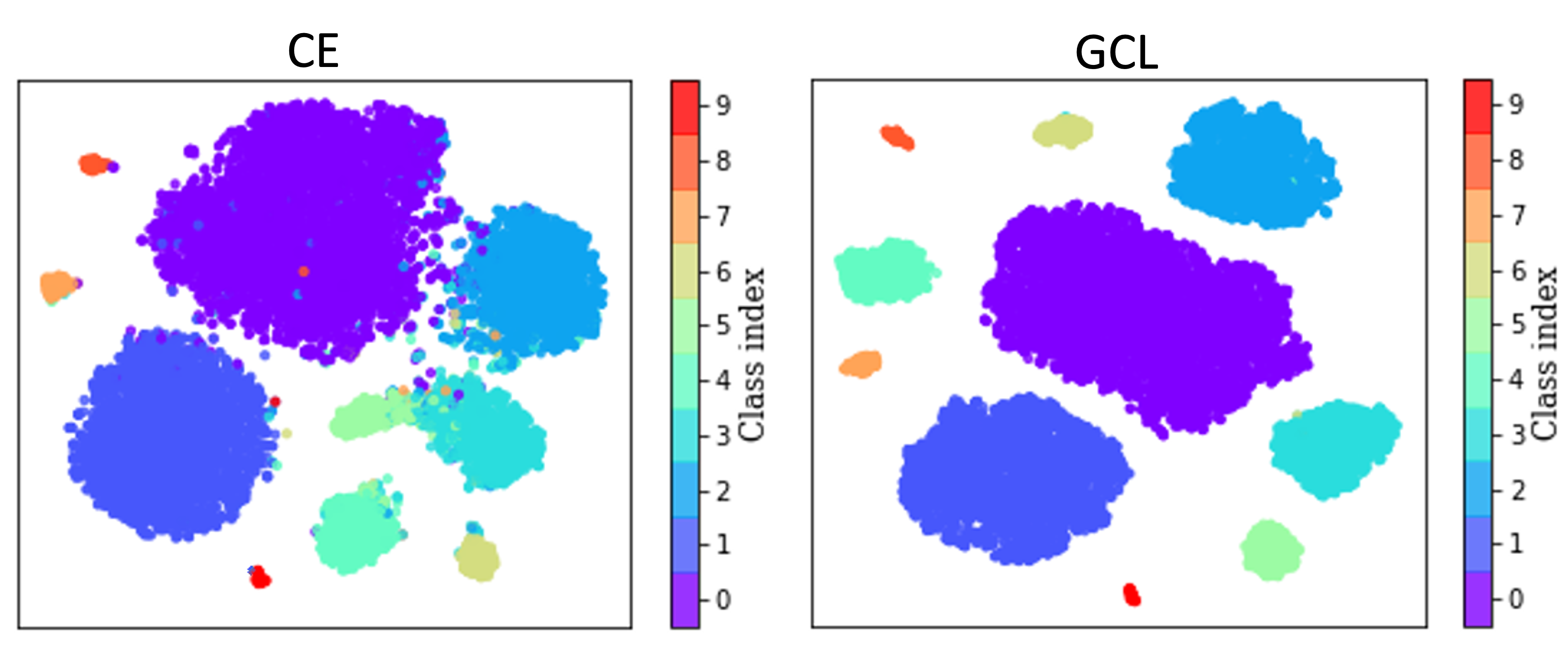}
        \label{fig:toy-train}}
  \\[-3pt]  
  \subfloat[On test set]{
        \includegraphics[width=0.48\textwidth]{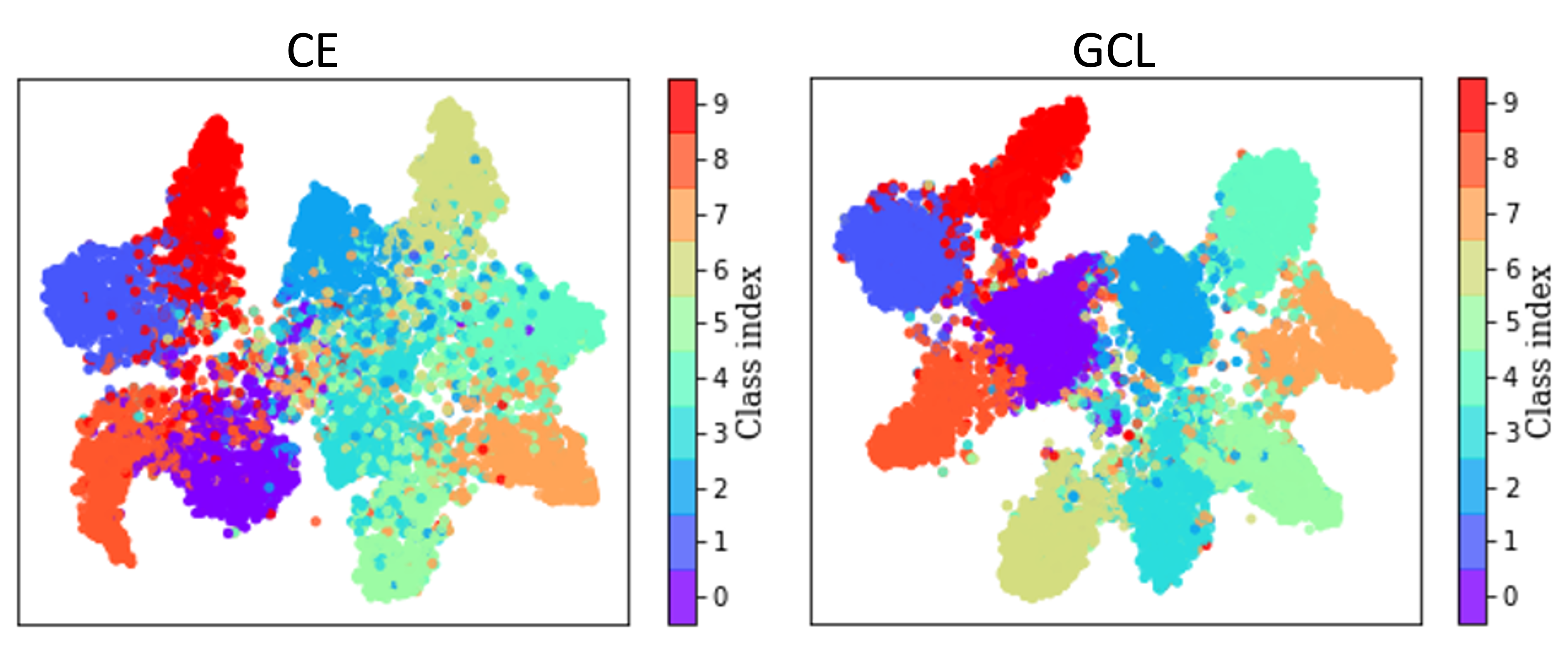}
        \label{fig:toy-test}}
\vspace{-6pt}
\caption{Visualization of the embedding via t-SNE from CIFAR-10-LT with $\gamma = 100$, where backbone network is ResNet-32. (Color for the best view.)}
\label{fig:visualization}
\end{figure}

In order to obtain additional insight, we utilized t-SNE projection of the embedding for visualization. Since the loss functions of baseline and MisLAS are both CE loss and MisLAS performed the second-best in most cases we have tried so far, we visualized CE loss embedding for comparison. The embeddings were calculated from the samples in CIFAR-10-LT with $\gamma = 100$. \cref{fig:visualization} shows the visualization results on the training and test set. From the result of the training set (\cref{fig:toy-train}), we can see that the embeddings obtained via GCL of different classes are more scattered. Therefore, the GCL embedding of each class is much easier to separate. The results of the test set shown in \cref{fig:toy-test} justify the efficacy of our proposed approach. The obscure region of CE loss embedding is larger than that of GCL embedding. Good embedding helps improve the model performance. We only re-fine the classifier with the simple cRT without any other complicated technologies, but the classification accuracy can be improved a lot.

\begin{table}[t]
\renewcommand{\thefootnote}{\fnsymbol{footnote}}
 \centering  
 \caption{Ablation experiment of different cloud size adjustment strategies (AS) on CIFAR-10-LT with $\gamma = 100$.} \label{ab_delta}
 \vspace{-6pt}
 \setlength{\tabcolsep}{10pt}
 \renewcommand{\arraystretch}{0.75}
  {\begin{tabular}{>{\centering}p{70pt}>{\centering}p{80pt}>{\raggedleft\arraybackslash}p{25pt}}
  \toprule[0.8pt]
  AS &Expression & Acc.(\%)  \\
  \hline
  \hline
  & &  \\[-6pt]
  cos.        &  $\cos (n_j/n_{max} \cdot \pi/2 ) $ & 79.21 \\
  & &  \\[-6pt]
  pow. diff. (e:1/3)  &$n_{max}^{1/3}-n_{j}^{1/3}$  &80.80 \\
  & &  \\[-6pt]
  pow. diff. (e:1/4) &$n_{max}^{1/4}-n_{j}^{1/4}$  &82.31 \\
  & &  \\[-6pt]
  log. diff.  &$\log n_{max}-\log n_{j}$   & \underline{\textbf{82.68}}  \\
  \bottomrule[0.8pt]
 \end{tabular}}
 \vspace{-6pt}
\end{table}

\subsubsection{Cloud Size Adjustment Strategy}\label{sec:cloud_size}
We explored several different cloud size adjustment strategies (AS), which included cosine form (cos.), power difference (pow. diff.) with different exponents (e:1/3 and e:1/4), and logarithmic difference (log. diff.). For a fair comparison, the sampler and re-training strategy were selected as CBEN and cRT, respectively. \cref{ab_delta} shows the results. We choose the log. diff. strategy according to \cref{ab_delta}.

\begin{table}[t]
 \begin{minipage}[t]{0.233\textwidth}
 \centering  
 \caption{Ablation experiment of different re-sampling strategy on CIFAR-10-LT with $\gamma = 100$.} \label{ab_rs}
 \vspace{-6pt}
 \setlength{\tabcolsep}{6pt}
 \renewcommand{\arraystretch}{0.75}
 {\begin{tabular}{c c c}  
  \toprule[0.8pt]
  Sam.&  RT & Acc.(\%)  \\
  \hline
  \hline
  & &\\[-6pt]
  IB  &cRT & 80.41 \\
  & &  \\[-6pt]
  CB  &cRT & 82.43 \\
  & &  \\[-6pt]
  EN  &cRT& 82.47 \\
  & &  \\[-6pt]
  CBEN &cRT & \underline{\textbf{82.68}}  \\
  \bottomrule[0.8pt]
 \end{tabular}}
 \end{minipage}
 \hfill
 \begin{minipage}[t]{0.233\textwidth}
 \centering  
 \caption{Ablation experiment of different re-training strategies on CIFAR-10-LT with $\gamma = 100$.} \label{ab_rt}
 \vspace{-6pt}
 \setlength{\tabcolsep}{4pt}
 \renewcommand{\arraystretch}{0.75}
  {\begin{tabular}{c c c }
  \toprule[0.8pt]
  Sam. &RT & Acc.(\%)  \\
  \hline
  \hline
  & &\\[-6pt]
  - &w/o RT   &80.52  \\
  & &  \\[-6pt]
  CBEN & LWS   &82.25 \\
  & &  \\[-6pt]
  CBEN & $\tau$-nor.  &82.16 \\
  & &  \\[-6pt]
  CBEN & cRT &\underline{\textbf{82.68}} \\
  \bottomrule[0.8pt]
 \end{tabular}}
 \end{minipage}
 \vspace{-6pt}
\end{table}

\subsubsection{Classifier Re-balance Strategies} \label{sec:sample_strategy}
We compared different strategies of data re-sampling and the classifier re-training to better analyze our proposed method. The re-sampling strategy (sam.) included: instance balance (IB)~\cite{decouple20}, class balance (CB)~\cite{decouple20}, class balance with effective number (EN)~\cite{cui2019class}, and our proposed class-based effective number (CBEN). For a fair comparison, the re-training strategies for all samplers were cRT. \cref{ab_rs} shows the effectiveness of CBEN.
For the selection of classifier re-training strategy, we first trained the backbone without any classifier re-training technology. Then, we fixed the representation and re-balance the classifier with learnable weight scaling (LWS)~\cite{decouple20}, $\tau$-normalization ($\tau$-nor.)~\cite{decouple20}, and cRT, respectively. \cref{ab_rt} presents the top-1 accuracy of CIFAR-10-LT with $\gamma = 100$. We can observe that, even without any classifier re-training technique, our approach can still beat most state-of-the-arts including two-stage methods. For example, our GCL without classifier re-training suppresses BBN by $0.7\%$. Further, cRT performs the best among the classifier re-training strategies, which improves the top-1 accuracy by $1.64\%$.   
From \cref{ab_rs} and \cref{ab_rt}, we can observe that IB+cRT degrades model performance, which indicates that training the classifier with IB may lead to classifier overfitting.

\section{Conclusion}
In this paper, we have found that softmax saturation reduces sample validity, which has different effects on head and tail classes. This implies that, from another perspective, softmax saturation can be utilized to automatically adjust the training sample validity of different classes. Subsequently, we have proposed the GCL. The tail class logits are set to relatively large cloud sizes to encourage more tail class samples to participate in training as well as leave large margins, which help obtain evenly distributed embedding space. The effectiveness of different classes is varied via GCL. Then, the simple but effective CBEN sampling strategy incorporated with cRT for classifier balancing has been proposed, which can further boost the model performance. Extensive experiments on various benchmark datasets have demonstrated that the proposed GCL has superior performance compared to the existing state-of-the-art methods.

\paragraph{Acknowledgment}This work was supported in part by NSFC/RGC JRS Grant: N\_HKBU214/21, ORP of Zhejiang Lab: 2021KB0AB03, GRF Grant: 12201321, NSFC Grants: 62002302 and 61672444, NSF of Fujian Province: 2020J01005, HKBU Grants: RC-FNRA-IG/18-19/SCI/03.
%
\clearpage
{\small
\bibliographystyle{ieee_fullname}
\bibliography{reference}
}

\end{document}